# High Accuracy Human Activity Monitoring using Neural network

Annapurna Sharma[1], Young-Dong Lee[2], Wan-Young Chung[3]
[1,2]Graduate School of Design & IT, Dongseo University, Busan 617-716, Korea
[3]Division of Electronics, Computer and Telecommunication Engineering, Pukyong National University, Busan 608-737, Korea
E-mail: [1]sharmaannapurna@gmail.com, [2]ydlee2@gmail.com , [3]wychung11@naver.com

## Abstract

*This paper presents the designing of a neural network for the classification of Human activity. A Tri-axial accelerometer sensor, housed in a chest worn sensor unit, has been used for capturing the acceleration of the movements associated. All the three axis acceleration data were collected at a base station PC via a CC2420 2.4GHz ISM band radio (zigbee wireless compliant), processed and classified using MATLAB. A neural network approach for classification was used with an eye on theoretical and empirical facts. The work shows a detailed description of the designing steps for the classification of human body acceleration data. A 4-layer back propagation neural network, with Levenberg-marquardt algorithm for training, showed best performance among the other neural network training algorithms.*

## 1. Introduction

Activity monitoring, in ubiquitous healthcare environment, gives an indication of user's activity level and hence can help in improving the quality of life. A recent study has shown that the number and share of the population aged 65 and over will continue to grow steadily over the next decades [1]. With less number of caregivers, activity monitoring system using compact, low-cost sensors can help in long term care with less expenditure. A number of approaches like vision system based and accelerometer based have been used in research since long [2][4][5][6]. For the classification purpose a decision tree, HMM, k-nearest neighbor classifiers and artificial neural network are the most popular approaches.

D.M. Karantonis et al. [6] presented a real time monitoring of human activity data received using a tri-axial accelerometer sensor. The algorithm classifies the activities such as rest, moving, falling and posture of the person (Standing, Lying sub postures and sitting). The system proposed majority of signal processing on the wearable unit. The classification was done using the separation of bodily and gravity acceleration components. The system also proposed an indirect measure of metabolic energy expenditure.

Juha Parakka et al. [2] showed their work on classification of daily activities like walking, running and cycling. The work shows the selection of sensors and requirement of Signal processing before the classification of the activities. Several time and frequency domain features were selected and classified using custom decision tree, automatically generated decision tree and artificial neural network.

In another work, Jonathan Lester et al [4], [5] presented a more detailed work using a shoulder mounted multi-sensor board. The sensor board was equipped to collect 7 different modalities. Three modalities were selected for activity recognition i.e. the audio, barometric pressure and accelerometer sensor. For the classification purpose, a total of 651 features were derived which included linear and Mel-scale FFT frequency coefficients, Cepstral coefficients, spectral entropy, band pass filter coefficients, integrals, mean and variances. A hybrid approach was used for recognizing activities, which combined boosting and learning an ensemble of static classifiers with Hidden Markov models (HMMs) to capture the temporal regularities and smoothness of activities. The work was shown to be able to identify ten different human activities with an accuracy of 95%.

This paper shows the complete designing of an artificial neural network for the classification of Human activity data received from an accelerometer sensor. The work gives a detailed description of designing the topologies of neural network, the selection of various training parameters. The features set was derived using the manual retrogression of data using only Fast Fourier transform (FFT) coefficients and hence the system uses only frequency domain attributes for classification. The use of compact, cheap wireless sensor unit along with the classification accuracy makes the system useful for monitoring the human activities ubiquitously.

## 2. Methods
### 2.1. Data Collection

The goal of the data collection was to assess the feasibility and accuracy of the activity classification using real data. The system composed of a chest worn wireless sensor unit and a base station unit. Telos type sensor node, TIP710 (MAX4, Korea) was used as a resource for computation and communication. Capacitive type Microelectromechanical Sensors (MEMS) tri-axial accelerometer MMA 7260 (Freescale Inc., USA) was used for capturing the acceleration signal of the movement. It has a range of -6g to 6g and sensitivity of 200mV/g, g is here the acceleration due to gravity in m/s$^2$. The software for sensor unit was developed using nes-C as a programming language with Tiny-OS [3][11]as the real time operating system to make it compatible with sensor network. TIP 710 has CC2420 2.4 GHz ISM band radio capable of data transfer at 250 kbps which transfers the three axis acceleration data to the base station sensor. The base station sensor sends the data to the PC for further processing.

### 2.2. Signal Processing and Feature extraction

The raw data, received at the base station PC, were acceleration data in terms of mV. Before processing the data it is required to convert it back to acceleration values. The process of conversion of the raw acceleration data to acceleration values in terms of 'g' is known as calibration. A linear calibration method was used so as to give the mean acceleration values of 0g, 0g and +1g for x, y, and z-axis respectively, when the user is vertical upright[9]. The same calibration gives the values of 0g, 0g and -1g for x, y and z-axis respectively, when the user is inverted vertical. The calibrated data is then smoothened using a moving average filter to remove any noise spikes present in the data. The data, so received consists of both body and gravity accelerations due to movement. For accurately monitoring the user activity and parameters involved, body and gravity acceleration data were separated. The gravity acceleration components lay within the low frequency ranges approx. 0-0.8 Hz [6]. A High pass elliptic filter was used to take out only body acceleration data. All the three axis Body accelerations were then combined to get a Root Mean Square (RMS) value of the three axis body accelerations,

$$RMS = \sqrt{X^2 + Y^2 + Z^2}$$

Where X Y and Z-are the calibrated body acceleration values of the corresponding axes in terms of 'g'. Figure 1 showing all the three axis acceleration data and the corresponding RMS acceleration data.

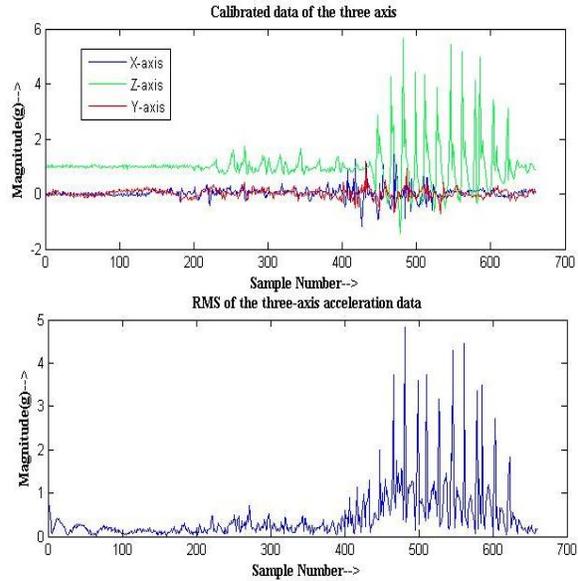

**Figure 1. The Calibrated data of the three axis accelerometer and corresponding RMS acceleration data**

The features for classification were calculated using the priori knowledge of frequencies associated with the walk and run activities. The Magnitude response during walking and running is shown in figure 2 and 3 respectively. The Frequency for walk lies in the range 1.5-2.5 Hz and that for Run lies in the range 2.5-5Hz [8]. Therefore a 128 point FFT of the RMS data is calculated. The choice of 128 is made so that the FFT coefficients were calculated for a power of 2 for fast computation [10] and also this value includes activity data of approximately 2.5 s, at a frequency of 50Hz [9]. This duration of activity data is sufficient to take the anomalies of human activities. Using the manual retrogression of data from the magnitude response figures, only initial 22 coefficients were used as a reduced feature vector which accounts to frequencies up to approximately 8 Hz [10]. A preprocessing of the extracted features is required to detect trends and flatten the distribution of the variable to assist the neural network in learning the relevant patterns [7] [12]. Therefore, the extracted features were so normalized that the mean will be zero and standard deviation be unity.

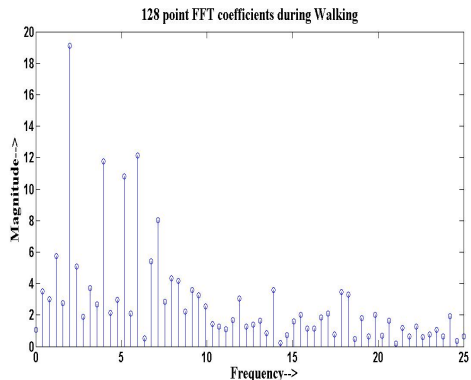

**Figure 2. Magnitude Response of RMS data during walking**

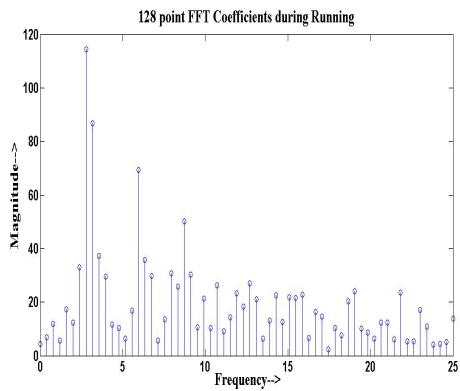

**Figure 3. Magnitude Response of RMS data during Running**

## 2.3. Training and testing data

The whole data so obtained is divided into training and test data sets. The training data was used for making the network learn from the data statics and hence setting the network weights and biases using a training algorithm. The testing data was used to check the performance of the neural network after training with the training data. The classification of testing data gives an indication of how well the network generalizes the classification for new data. A 3-fold cross validation method was used in which the whole data was divided into 3-folds: each of which composed of equal number of rest, walk and run data sets. Out of the three folds, two were used as a training data sets and the rest as the test data. The training and testing process was repeated for all possible combinations of the three folds.

## 2.4. Neural Network Paradigms

This section discusses the designing of the back propagation neural network classifier for human activity classification. A Back propagation neural network consists of a collection of inputs and processing units known as neurons. The neurons are arranged into layers: Input layer, Hidden layer and Output layer. A general architecture of an artificial neural network is shown in figure 4. The neurons in each layer are fully interconnected by connection strengths called as weights. Also each hidden and output layer neuron consists of a bias term associated with it. The designing of neural network involves the decision in number of hidden layers and number of neurons in each layer along with the transfer functions. In this work MATLAB neural network toolbox has been used for implementation.

**2.4.1. Number of input neurons.** The number of input neurons was selected as 22 because the number of independent variables in the preprocessed feature data was chosen as the same.

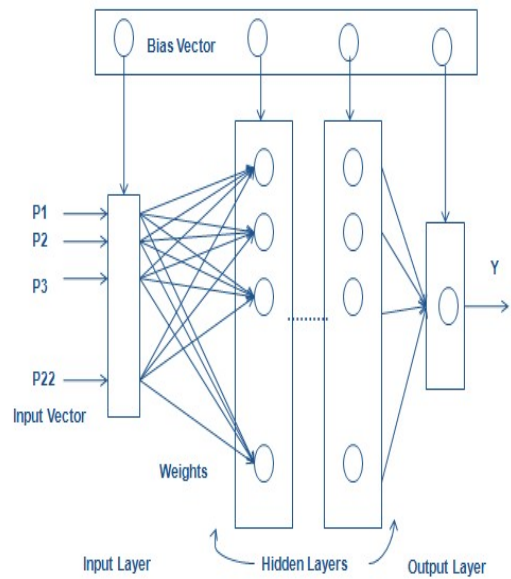

**Figure 4. Architecture of a back propagation neural network**

**2.4.2. Number of output neurons.** This depends upon the number of class labels desired as well as encoding scheme used. We want to classify the data in one of the three classes i.e. Rest, Walk and Run. Using the binary encoding would require an equal number of output neurons as the class labels. Also, if more activities were to include, it will take more variables and hence more number of neurons with each activity. So a linear transfer function is used for the output neuron with a decimal encoding. This requires only one neuron for output layer. The three classes were encoded as '0' for the Rest, "1' for the Walk and '2' for the Run activity.

**2.4.3 Transfer function.** Transfer function of any layer determines its output. A sigmoid function 'tansig' was used as a transfer function for each hidden layer and linear function 'purelin' for the output layer [12]. The data has already been normalized with mean =0 and standard deviation=1, so that it is consistent with the transfer function. The sigmoid function is a nonlinear and continuously differentiable function. It can convert an infinite range into a finite one and hence prevent outputs from reaching very large values which can inhibit the training process. The 'purelin' function for output layer gives ease in further increasing the number of output classes without changing the number of neurons because a single neuron can give different number of classes with decimal encoding [12].

**2.4.4 Error function.** The error function defines the evaluation criterion. We chose a mean squared error function 'mse' as an evaluation criterion. It minimizes the mean of the squares of the errors produced in each iteration and updates the network weights and biases accordingly. The performance goal for the training is decided using this error function only. The value of goal parameter for 'mse' is shown in table 3.

**2.4.5. Number of hidden layers and hidden neurons.** The selection of number of Hidden layer and neurons in each layer were determined empirically. Initially the network was trained and tested with one hidden layer and then by increasing the number of neurons. The classification results are shown in table 1. The table shows that increasing the number of Hidden layer neurons results in increase in mean classification rate. The mean of 10 runs for training and testing has been recorded because sometimes the training got struck at 33% due to random weight initialization. The table also shows that seven numbers of neurons are giving the best classification rate for both the testing and training data sets with relatively less standard deviation. So the number of neurons, for the first layer, was fixed as 7. After deciding the first hidden layer, a second hidden layer was then added to the network to check for any accuracy improvement for classification. The second hidden layer was then trained and tested with an increasing number of neurons. The mean classification rate and standard deviation for both the train and test data are included in table 2. A comparison of tables implies that with the increase of number of hidden layers the accuracy is improved and the standard deviation is decreased. Also the mean test data classification rate is higher for 7 numbers of neurons in hidden layer 2. Therefore, the overall network was fixed to have N1=7 and N2=7. After this, a third layer was added and checked for performance but the improvement in classification rate was not so significant that the final design of neural network was fixed to have just two hidden layers.

**Table 1. Classification accuracy with changing the number of neurons in Hidden Layer 1**

|  | Number of neurons in Hidden layer (N1) | | | |
| --- | --- | --- | --- | --- |
|  | 3 | 5 | 7 | 9 |
| Mean Training Classification Rate % | 55.44 | 74.52 | 78.56 | 73.51 |
| Training Standard Deviation | 39.85 | 28.63 | 29.57 | 29.53 |
| Mean Testing Classification Rate % | 52.89 | 68.86 | 77.08 | 77.08 |
| Testing Standard Deviation | 30.89 | 27.11 | 7.65 | 12.88 |

**Table 2. Classification accuracy with changing the number of neurons in Hidden Layer 2**

|  | Number of neurons in Hidden layer 2 (N2) with N1=7 | | | |
| --- | --- | --- | --- | --- |
|  | 3 | 5 | 7 | 8 |
| Mean Training Classification Rate % | 87.17 | 85.56 | 91.82 | 92.32 |
| Training Standard Deviation | 0.1804 | .2255 | 0.047 | 0.0502 |
| Mean Testing Classification Rate % | 78.13 | 79.37 | 83.96 | 78.75 |
| Testing Standard Deviation | 0.1905 | .2316 | 0.118 | .1190 |

## 2.5. Neural Network Training

The objective of the neural network training was to find the weights and biases between the neurons that determine the global minimum of the error function. We have used Levenberg-marquardt algorithm to train the classifier for the classification of Human activity data from the accelerometer sensor. For the activity recognition, a fast training was supposed to use.

Therefore, the Levenberg-marquardt algorithm is used for training the neural network. Finding the global minimum also requires selection of some parameters. MATLAB provides the default values for these parameters. An empirical approach was used to decide for modifying the parameter values for the designing of the neural network. The values of parameter used in our design are shown in table 3 with a brief description of each of the parameter [12].

**Table 3. Parameters used for neural network training**

| Parameter | Default value | Description |
|---|---|---|
| epochs | 7500 | Maximum number of epochs to train |
| goal | 0.01 | Performance goal |
| max_fail | 5 | Maximum validation failures |
| mem_reduc | 1 | Factor to use for memory/speed trade off |
| min_grad | 1e-10 | Minimum performance gradient |
| mu | 0.0001 | Initial Mu |
| mu_dec | 0.1 | Mu decrease factor |
| mu_inc | 10 | Mu increase factor |
| mu_max | 1e10 | Maximum Mu |
| show | 25 | Epochs between displays (NaN for no displays) |
| time | inf | Maximum time to train in seconds |

## 3. Experimental Results and discussion

The performance of the neural network is evaluated with the increasing number of hidden layers as well as each hidden layer neurons for both the training and testing data. With respect to the training data, the neural network is evaluated for over fitting i.e. if the classification rate is 100%, in such a case the network may not be able to generalize the classification for other data. Also the testing data classification rate signifies how well the network can classify the data which were not viewed during the training. The classification results, for the data at hand, are shown in table 1 and 2. A complete analysis of the result was done for deciding the number of hidden layers and the number of neurons in each layer in section 2.4. The highest mean classification rate achieved using 3-layer neural network (i.e. one hidden layer) is 77.08%. The highest mean classification rate with the four layer neural network (i.e. 2 hidden layers) is 83.96% for the testing data.

**Table 4. Comparison of the neural network classifier with other approaches**

| Approach for classification | | Mean Training classification rate % | Mean Testing classification rate % |
|---|---|---|---|
| With Neural network classifier | One Hidden layer (N1=7) | 78.56 | 77.08 |
| | Two Hidden layer (N1=7, N2=7) | 91.82 | 83.96 |
| Without Neural network classifier | | - | 75.00 |

As compared to other methods that do not use neural network classifier and consider only the FFT of the acceleration data [8], the neural network classifier is giving a better classification rate. A comparison of the classification rate using these different methods is shown in Table 4. A comparison shows that the mean classification rate with the neural network classifier is more as compared the previous results that were computed without the neural network classifier but the same signal processing and classification features. Also adding another hidden layer with 7 neurons further improves the mean classification rate.

## 4. Conclusions

A neural network classifier has been designed for human activity data. The classification is done using only frequency domain feature. A fast training algorithm i.e. Levenberg-marquardt algorithm was used for training. The designed network (with 2 hidden layers) is giving a mean classification rate of 91.82 + 0.047% and 83.96 + 0.118 % for training and test data sets respectively. The mean classification rate with neural network classifier is also an improved one as compared to the previous results without the neural network classifier. The classification accuracy and use of wireless sensor node makes the system compatible to use in ubiquitous home and health environment.